\documentclass[10pt,twocolumn,letterpaper]{article}

\usepackage[pagenumbers]{cvpr} 

\usepackage{graphicx}
\usepackage{amsmath}
\usepackage{amssymb}
\usepackage{booktabs}

\usepackage{color}

\definecolor{ben}{rgb}{0.9,0.,0.5}

\definecolor{mahdi}{rgb}{0.2,0.5,0.2}

\usepackage[pagebackref,breaklinks,colorlinks]{hyperref}

\usepackage[capitalize]{cleveref}
\crefname{section}{Sec.}{Secs.}
\Crefname{section}{Section}{Sections}
\Crefname{table}{Table}{Tables}
\crefname{table}{Tab.}{Tabs.}

\usepackage{makecell}
\usepackage{multirow}
\usepackage{bm}

\usepackage[symbol]{footmisc}
\renewcommand\footnotemark{}

\def\cvprPaperID{1216} 
\def\confName{CVPR}
\def\confYear{2022}

\begin{document}

\title{ZebraPose: Coarse to Fine Surface Encoding for 6DoF Object Pose Estimation}

\author{
\begin{tabular}{cccc}
Yongzhi Su\textsuperscript{1,2*}&
Mahdi Saleh\textsuperscript{3*}&
Torben Fetzer\textsuperscript{2}&
Jason Rambach\textsuperscript{1}
\\
Nassir Navab\textsuperscript{3} &
Benjamin Busam\textsuperscript{3} &
Didier Stricker\textsuperscript{1,2} &
Federico Tombari\textsuperscript{3,4} \vspace{0.2cm}
\end{tabular}
\\
\textsuperscript{1} German Research Center for Artificial Intelligence (DFKI)\hspace{1.5cm}
\textsuperscript{2} TU Kaiserslautern\\
\textsuperscript{3}Technische Universit\"{a}t M\"{u}nchen \hspace{1.5cm} 
\textsuperscript{4}Google\\
\tt\small{\{yongzhi.su; jason.rambach; torben.fetzer; didier.stricker\}@dfki.de}\\
\tt\small{\{m.saleh; b.busam; nassir.navab\}@tum.de}, \tt\small{tombari@in.tum.de} 

\thanks{\textsuperscript{*}The authors contributed equally to this paper}
\thanks{Code: \href{https://github.com/suyz526/ZebraPose}{https://github.com/suyz526/ZebraPose}}
}
\maketitle


\begin{abstract}
Establishing correspondences from image to 3D has been a key task of 6DoF object pose estimation for a long time. To predict pose more accurately, deeply learned dense maps replaced sparse templates. Dense methods also improved pose estimation in the presence of occlusion. More recently researchers have shown improvements by learning object fragments as segmentation. In this work, we present a discrete descriptor, which can represent the object surface densely. By incorporating a hierarchical binary grouping, we can encode the object surface very efficiently. Moreover, we propose a coarse to fine training strategy, which enables fine-grained correspondence prediction. Finally, by matching predicted codes with object surface and using a PnP solver, we estimate the 6DoF pose. Results on the public LM-O and YCB-V datasets show major improvement over the state of the art w.r.t. ADD(-S) metric, even surpassing RGB-D based methods in some cases.
\end{abstract}


\section{Introduction}
\label{sec:intro}

Augmented reality and robotics are two of the main application fields of 3D computer vision. In many augmented reality applications, the location and pose of an object of interest has to be determined at a high precision~\cite{marchand2016pose, rambach20186dof}. Similarly, object grasping and manipulation is needed for many robotic applications (e.g. automatic manufacturing~\cite{perez2016robot},  cooperative assistance~\cite{busam2015stereo,ghazaei2018dealing}), and also demands accurate 6 Degree-of-Freedom (6DoF) object pose information. As the crucial element in both application domains, estimating the 6DoF object pose has received increasing attention from the computer vision research community.

\begin{figure}
\centering
\includegraphics[width=\linewidth]{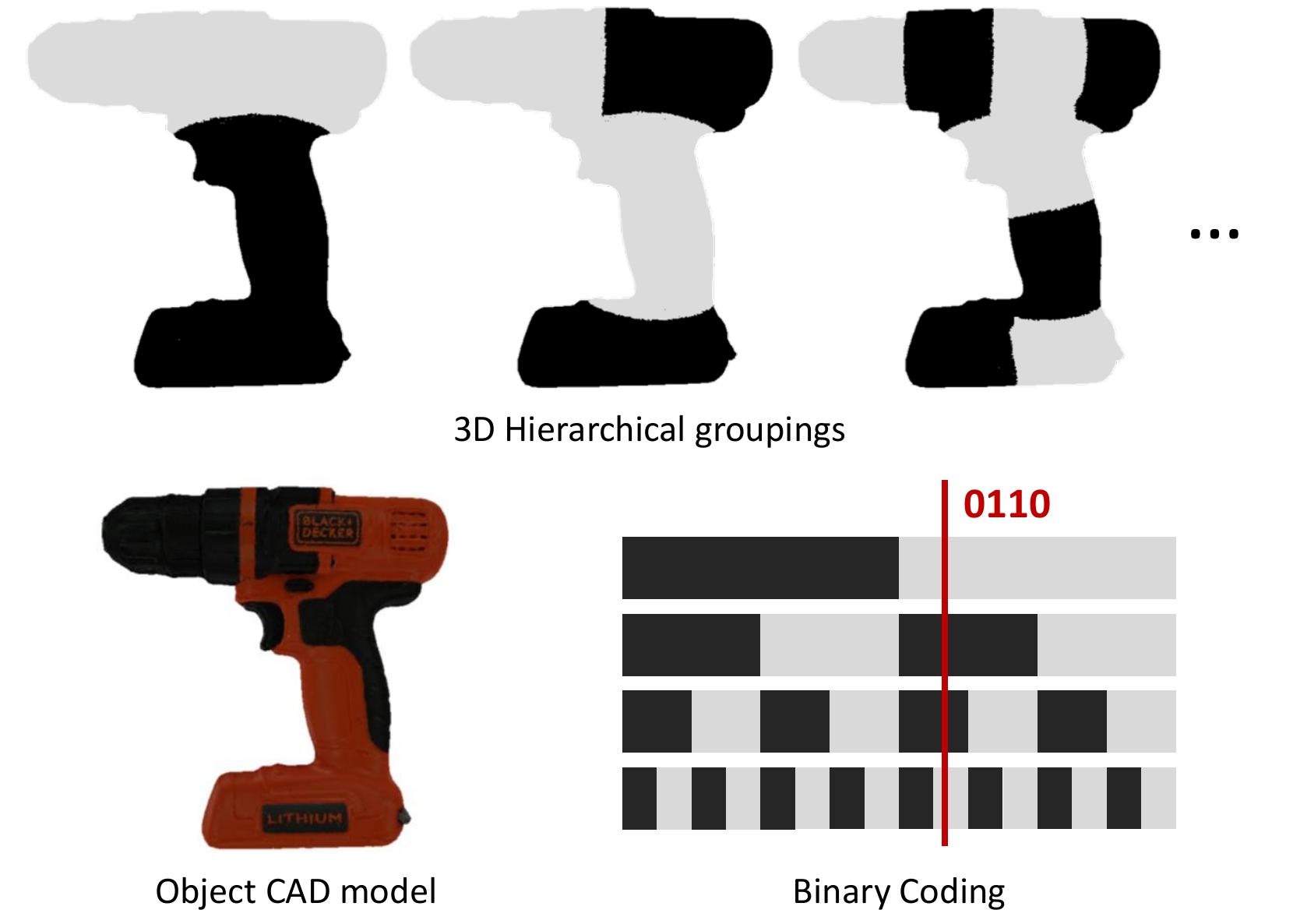}
\caption{ZebraPose assigns a discrete code to each surface vertex hierarchically. We project the code as binary black and white values (top) and learn them using deep neural networks. Our binary descriptor allows one-to-one correspondence for the problem of 6DoF object pose efficiently.  }
\label{fig:teaser}
\end{figure}

The correspondence problem is a classical problem in computer vision. While finding correspondences across the same domain is more straightforward, estimating the 6DoF object pose requires 2D-3D correspondences. In earlier object pose estimation research, depth maps came to help to match image pixels to 3D surface points~\cite{hinterstoisser2012, vidal20186d}. Due to cost and setup complications, the detection of 6DoF pose without depth information can be advantageous. However, RGB approaches typically achieve a lower accuracy with respect to their depth-based counterparts~\cite{collet2011moped, hinterstoisser2012gradient}.

Driven by the recent developments in deep learning and Convolutional Neural Networks (CNNs), various methods were proposed, which make 6DoF pose estimation from a single RGB image feasible~\cite{sundermeyer2018implicit,kehl2017ssd,xiang2017posecnn,busam2020like, rambach2018learning}. In a correspondence-based setting, to estimate the object pose, Perspective-n-Points (PnP) algorithms require at least 4 2D-3D point matches~\cite{epnp_2009}. Therefore, sparse methods are applied to extract points of interest \cite{rad2017bb8,raddeepheatmaps}. However such methods might fail to find object landmarks under viewpoint changes, occlusion, or lack of texture.
With the success of deep neural networks in image synthesis problems, researchers use such tools to generate dense correspondence maps. For instance, several methods learn UV~\cite{Zakharov2019DPOD6P} or UVW ~\cite{wang2019normalized,park2019pix2pose} values in object local coordinates. Since the network produces dense smooth results, certain low-level geometry is lost. Moreover, neural networks tend to achieve a higher performance in classification tasks \cite{kehl2017ssd}. 

To this aim, we propose a dense correspondence pipeline that combines the concepts of handcrafted features and image segmentation in a hierarchical fashion for RGB-based 6DoF pose estimation. 
In order to design a descriptor that encodes surfaces efficiently, we use the binary numeral system. Binary-based descriptors are applied in ORB ~\cite{orb_2011} and are still in use in robust SLAM applications~\cite{campos2021orb}. In our work, we split the surface into halves in multiple iterations and define our vertex encoding by stacking the assigned group labels. By leveraging a hierarchical discrete representation, we guarantee a compact mapping and simple learning objective as a multi-label classification problem \cite{kehl2017ssd,hodan2020epos}.
Moreover, learning how to encode a full sequence at once might be challenging for neural networks. Therefore, we propose a coarse to fine learning scheme. By design, our encodings on the coarse levels are continuously shared in wider object regions. As the network learns to differentiate coarse splits, we focus on finer encoding positions. With a coarse to fine loss and training strategy, we then manage to predict fine-grained surface correspondences. 

In contrast to previous works where there is no guaranteed putative correspondence~\cite{park2019pix2pose,peng2019pvnet,wang2019normalized}, our encoding promotes direct pixel-to-surface matching just by means of a look-up table. With a simple matching and PnP-RANSAC scheme of Progressive-X~\cite{barath2019progressive}, we outperform the state of the art in 6DoF pose on the most commonly used benchmarks w.r.t.  ADD(-S) metric.

In summary, we propose ZebraPose, a two-stage RGB-based approach that defines the matching of dense 2D-3D correspondence as a hierarchical classification task. We divide the general two-stage approach for 6DoF object pose estimation into three components: 1) assigning a unique descriptor to the 3D vertex; 2) predicting a dense correspondence between the 2D pixels and 3D vertices; 3) solving the object pose using the predicted correspondences. We can summarize our proposed contributions in this paper related to the first two components:
\begin{itemize}
\item A novel coarse to fine surface encoding method assigning the dense vertex descriptor in an efficient way, which also fully exploits traditional outlier filters used in computer vision task. 
\item A novel hierarchical training loss and strategy to automatically adjust the weights of each code position.   
\end{itemize}
Extensive experiments on LM-O~\cite{brachmann2014} and YCB-V~\cite{calli2015ycb} datasets show that our proposed approach achieves state of the art results. 

\section{Related Work}
We limit our in-depth discussion of related work to the most relevant methods to our work, i.e. RGB-based 6DoF pose estimation, and object surface encoding techniques.

\subsection{RGB-based 6DoF Pose Estimation}
\textbf{Traditional Methods.}
With the development of the feature descriptor~\cite{lowe2004distinctive}, the object pose problem could be solved by feeding estimated 2D-3D correspondences into a RANSAC/PnP framework. However, dealing with texture-less objects remained a challenge. To overcome the lack of keypoints, Hinterstoisser \etal~\cite{hinterstoisser2011gradient} proposed to utilize the image gradient information and formulate the pose estimation task within a template matching pipeline. Later advances~\cite{brachmann2016uncertainty} avoided the template searching time by applying a statistical learning-based framework to regress object coordinates and object labels jointly. 
However, the accuracy that handcrafted methods can achieve is far from that of deep learning methods nowadays.

\textbf{End-to-End Methods.}
PoseNet~\cite{kendall2015posenet} was the first work that attempted to regress the camera viewpoint with a CNN. Following works usually concatenated an object detector with the pose regression, making multi-object pose estimation possible~\cite{xiang2017posecnn}. Finding a suitable rotation representation for pose regression was a problem at that time and typical rotation parametrization did not populate Euclidean spaces~\cite{busam2017camera}. SSD6D~\cite{kehl2017ssd} avoided complex parameters by discretization of the rotation space thus treating the rotation estimation as a classification problem. Zhou et.al~\cite{zhou2019continuity} proposed a continuous 6-dimensional rotation representation that shows advantages over quaternions~\cite{manhardt2018deep,manhardt2019explaining} or Lie algebra~\cite{do2019real,su2021synpo} parametrization for neural network training. This representation is utilized in several direct regression works~\cite{labbe2020cosypose,wang2021gdr,di2021so}.

In parallel, several efforts have been made to integrate RANSAC and PnP modules to pose learning frameworks. ~\cite{brachmann2018learning,brachmann2019neural,brachmann2017dsac} propose differentiable RANSAC variants, which are not applicable to object pose estimation as they require a good initialization and complex training strategy.\cite{hu2020single} proposes a network to solve the PnP problem, with a loss function reflecting pose metrics. At the same time, a new branch of methods has been developed with the growth of neural renderers~\cite{loper2014opendr,kato2018neural,chen2019learning}. \cite{iwase2021repose} is able to define the loss according to the texture colour on pixel level. \cite{wang2020self6d,sock2020introducing} used a differentiable depth map and achieved self-supervised network fine-tuning with unlabeled RGB-D data. In an effort to combine correspondence-based methods with direct regression of 6DoF parameters, \cite{wang2021gdr} used correspondence maps as an intermediate geometric representation to regress the pose. \cite{di2021so} further enhances \cite{wang2021gdr} by employing self-occlusion information that provides richer information to predict the object pose with the predicted 2D-3D correspondences. 

\textbf{Indirect Methods with Deep Learning.} While end-to-end methods have evolved through time by integrating differentiable modules, the performance of such methods are normally below geometrical and indirect methods. Combining learning features and geometrical fitting, 
\cite{wohlhart2015learning} uses metric learning to learn an implicit pose representation through triplet loss and finally looks for nearest neighbors in pose space. AAE~\cite{sundermeyer2018implicit} 
learns to generate a latent vector based on the visual information of the object in discrete viewpoints. At inference stage, the rotation is obtained by comparing the latent code with the pre-generated rotation-latent code lookup table. The rest of the indirect methods usually estimate the 2D-3D correspondence, and solve the object pose using RANSAC/PnP. BB8~\cite{rad2017bb8} firstly defines the 3D object bounding box corners as the keypoints and PVNet~\cite{peng2019pvnet} reaches high recall rate in LM~\cite{hinterstoisser2011multimodal} dataset by predicting the keypoints with a dense pixel-wise voting for sampled keypoints on the object. The main drawback of such sparse 2D-3D correspondence methods is that the prediction of keypoints in the occluded area lacks in accuracy. HybridPose~\cite{song2020hybridpose} proposed to leverage multiple geometric information to tackle this issue while other methods~\cite{park2019pix2pose,Zakharov2019DPOD6P,hodan2020epos} predict pixel-wise dense 2D-3D correspondences.

\subsection{Surface Encoding}
The binary surface encoding technique has been successfully used in the field of structured light reconstruction for many years~\cite{mimou1981method,posdamer1982surface,trobina1995error,salvi2004pattern}. For this purpose, a video projector illuminates the scene with several successively refined binary fringe patterns. 
The composition of the different stripe patterns provides an encoding of the surface points. 
Surface coding using multiple classification problems 
has proven to be highly reliable and competitive \cite{giancola2018survey}. Since neural networks are ideally suited for solving classification problems, a transfer of the approach as we presented in this work constitutes a logical step. 

In pose estimation domain, to estimate the dense 2D-3D correspondence, each 3D corresponding point must be assigned a unique descriptor. Pix2Pose~\cite{park2019pix2pose} simply treats the 3D vertex coordinates as this descriptor. DPOD~\cite{Zakharov2019DPOD6P} textures the object with a 2-channel UV-map with discrete values 
to learn the correspondences. EPOS~\cite{hodan2020epos} divides the object surface into multiple fragments, and estimates the corresponding points by combining fragment segmentation and local fragments coordinates prediction. Although most of these encodings are limited to local object coordinates, we propose a method that learns dense 2D-3D correspondence through a handcrafted code. Compared to methods that predict local coordinates space ~\cite{Zakharov2019DPOD6P,wang2019normalized} in 2D or 3D grid, we encode the object surface in a coarse to fine manner. Moreover, unlike EPOS~\cite{hodan2020epos} that divides the object surface into multiple coarse bins at once, we divide the object surface iteratively until the fragments are fine enough to define the unique 3D corresponding point. This allows for gradual refinement of the correspondence through the hierarchical levels.


\section{Method: ZebraPose} 

\begin{figure*}[t]
\centering
\includegraphics[width=\textwidth]{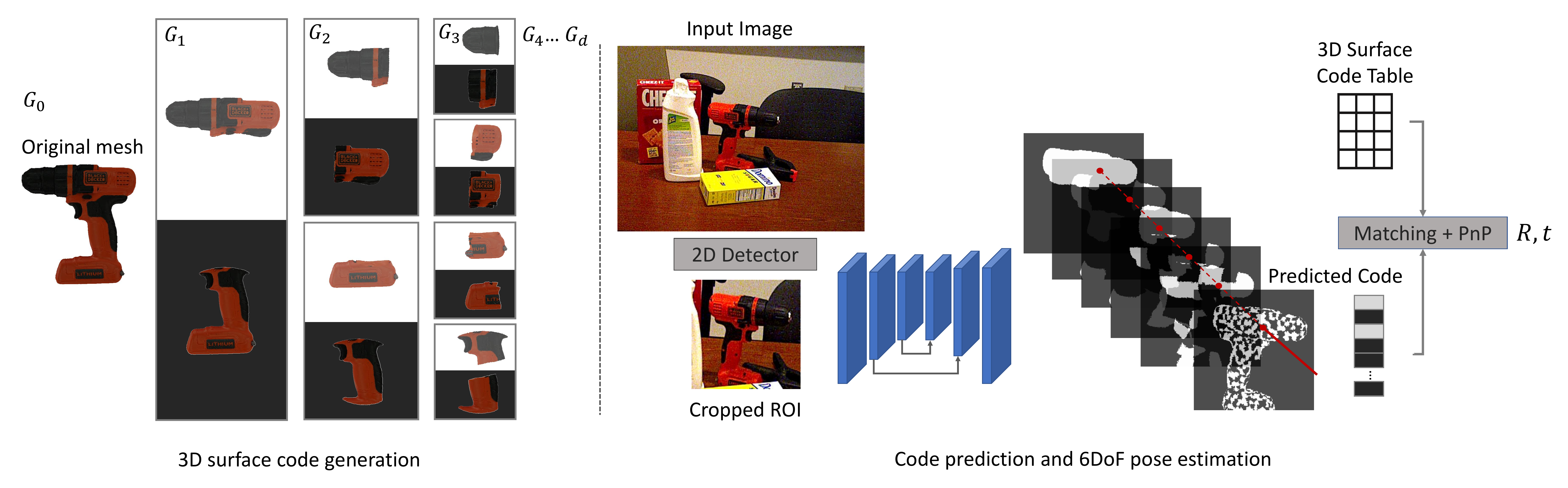}
\caption{Left: Our hierarchical encoding is defined by grouping surface vertices in several iterations. In each iteration, object vertices are split into equally sized groups. In a binary setting, vertices are classified into two groups, 0 (white) and 1 (black). This process happens offline and the generated mapping between vertex code and the corresponding 3D vertex is stored in a look-up table. Right: Our training framework uses a detector to crop the object ROI and predicts a multi-layer code using a fully convolutional neural network. The predicted code is then matched to the 3D surface vertex and passed to RANSAC and PnP modules for pose estimation. }
\label{fig:overview}
\end{figure*}

In this section, we present our approach for the problem of 6DoF object pose, which involves the entire process from our surface encoding to the final pose estimation. 


\subsection{Coarse to Fine Surface Encoding} \label{sec:SurfaceEncoding}
Given a surface CAD model of an object and its vertices $\bm{v_i} \in \mathbb{R}^3$ , where $i$ stands for the vertex id, we want to represent each $\bm{v_i}$ with vertex code $c_i \in \mathbb{N}^d$, where $d$ is the length of the vertex code. We need to define such encoding based on vertices' position relative to the given 3D object surface to enable coarse to fine learning. To enable this, we construct our codes in a non-decimal numeral system.

Defining our encoding in a numeral system with lower radix makes the representation very efficient and provides easier grounds for coarse to fine grouping of the points. For a code of length $d$, we perform $d$ iterations of grouping of the vertices. The collection of groups $G_j$ in the $j$-th iteration ($j \in \{0,...,d\}\subset\mathbb{N}$) consist of $r^j$ groups. $G_0$ defines the initial group, including only one group, i.e. the entire object vertices. $G_j$ with $j > 1$ is obtained by splitting each group in $G_{j-1}$ into $r$ groups. In a grouping iteration, each vertex $\bm{v_i}$ is assigned with a class id $m_{i,j}$, where $m_{i,j} \in \{0,...,r-1\} \subset\mathbb{N}$ based on the group it belongs to in the $j$-th grouping. Finally, each vertex is assigned to a vertex code with $d$ digits by stacking the class id of each grouping operation. This representation is stored and fixed for every 3D object. The vertices in each group share the same code. We build the lookup table to map a code to the centroid of the each group in $G_d$, which is further used to build 2D-3D correspondence and solve the pose as described in Sec.~\ref{pose}. In this paper, we used k-means for the grouping, more details are in Sec.~\ref{sec:Experiments_Setup}. We illustrate this process in Fig.~\ref{fig:overview} with $r = 2$ and break down the CAD model surface into discrete and equally sized groups. 

\subsection{Choice of the Radix for Vertex Code}\label{sec:choice_of_vertex_code}
Following our grouping described in Sec.~\ref{sec:SurfaceEncoding}, we would have $K$ total number of classes, where $K=r^d$. In a classification problem we learn these maps using $o$ logits, where $o=r \cdot d$.  To minimize the number of outputs while learning the most number of classes we have:
\begin{equation}
    o_{min} = \min_r r \cdot d =  \min_r r \cdot log_r K =  2 \cdot log_2 K.
\end{equation}
The best positive integer choice of $r$ to minimize the number of network layers are 2 and 4. Since a value is classified either as positive or negative, we do not need to use the cross entropy loss with 2 explicit output layers for the binary classification. So we can reach $log_2 K$ as the optimal number of output layers with $r=2$. 

Besides the advantages of reduced GPU memory requirement, we show later in the ablation study (see Sec.~\ref{sec:ablation_study}) that using the binary vertex code yields the most accurately predicted pose. Thus we choose a binary base for the vertex code.

\subsection{Rendering the Training Labels}\label{sec:render_training_labels}
Each object pixel in the image corresponds to a 3D object vertex.
The network predicts the class id that is assigned to this vertex in each grouping operation. Therefore, we still need to render the class id into the 2D image plane with a given pose for the training. For this purpose, we transfer the class id of vertices into the class id of the mesh faces using the following criteria: if two vertices of a face have the same class id, the face is assigned with this class id. Otherwise, the face has the class id of its first vertex. We repeat this rendering process for $d$ times until the training label class id for each grouping is generated.


\subsection{Network Architecture} \label{NetworkArchitecture}
In Sec.~\ref{sec:choice_of_vertex_code} we justify our choice of $r=2$. In this regard, our goal is to classify $2^d$ regions with only $d$ binary values. 

During training, we use the object pose annotations to render the labels as layered black and white maps to image coordinates. This way, our objective learning maps are $d+1$ binary labels ($d$ for the binary vertex code and 1 for the object mask) for code and visible mask prediction. An encoder-decoder network generates $d+1$ outputs with a single decoder. We round the final output probabilities to represent our discrete vertex codes.

The entire process from input images to the predicted pose is presented in Fig.~\ref{fig:overview}. To predict the code per pixel in the frame with fine granularity, we process only a Region of Interest (ROI) around object pixels. Following the pipeline similar to \cite{wang2021gdr,li2019cdpn,labbe2020cosypose}, we focus on the object pose and use the available 2D detector predictions to find the ROIs. We crop and resize the ROI from the prediction to a fixed dimension H$\times$W, and apply the exact process to the target vertex code maps during training. Our goal is to predict multiple labels per frame in the ROI. 


\subsection{Hierarchical Learning} \label{Learning}
Predicting correspondences directly from object pixels is a fine-grained task. On the other side, deep neural networks are commonly used for coarse level predictions. This means features predicted per pixel are very similar in a small vicinity. As our encoding is also hierarchical by design, we learn the codes in a coarse to fine manner. Therefore, the predictions are learned in different stages, from coarse groupings to fine ones. We use an error histogram for each position on hierarchical level and weight our Hamming-based loss given the error to design this.

\textbf{Mask loss.} Firstly, we predict the visible mask to segment the object area from the background. Here, we simply pass the predicted probability to the sigmoid function and use L1 loss as $L_{mask}$. It is worth noting, for the binary vertex code prediction in the following, we only calculate loss of the pixels within the predicted object mask.

\textbf{Hamming distance:} The CNN outputs the binary vertex code probabilities $\hat{p} \in \mathbb{R}^d$ for a pixel within a ROI, we obtain the predicted discrete binary code $\hat{b}$ by rounding $\hat{p}$. Given $\hat{b}$  and its known ground truth binary vertex code $b$, the Hamming distance \text{$Hamm$} is defined by counting the number of bits $\hat{b}$ which are different from $b$. This formulation does not favor any of the positions and calculates the error without considering any hierarchical information explicitly.
As a common practice in deep learning, we use binary cross-entropy as an activation function for the distance:
\begin{equation}
    \text{Hamm}(b,\hat{p}) = \sum_{j=1}^{d}{b_j} \log \hat{p_j} +(1-b_j)\log (1-\hat{p_j}),
\end{equation}
where $b_j$ stands for the $j$-th bit in $b$ (the $j$-th bit is generated in the $j$-th vertices grouping).

\textbf{Active bits.} Lower bits in binary vertex code $b$ hold coarse correspondences, and higher bits define finer estimates. During the initial training phase, the network focuses on learning the coarse splits and has a higher error on fine bits. Therefore we adaptively weight the coarse bits by looking at the histogram of error of all bits. As the training proceeds and coarser predictions become more robust, finer bits are induced with more weights. We define our histogram at training step $t$ by looking at the error at different bits:
\begin{equation}
    H_j(t)= avg(\lambda (b_j^t-\hat{b_j}^t) + (1-\lambda) (b_j^{t-1}-\hat{b_j}^{t-1})),
\end{equation}
where $\hat{b_j}^t$ defines the predicted binary vertex code $\hat{b_j}$ at training step $t$, and $\lambda$ is a constant. With the $avg$ operator, we get the error ratio by calculating the average difference in $b_j^t$ and $\hat{b_j}^t$ of all pixels within the predicted object mask in a mini-batch. 
During training we update the histogram given the previous histogram in training step $t-1$ and the current error histogram. We show how to define a hierarchical loss based on the histogram in the following.

\textbf{Hierarchical loss.}
We compute a weighting coefficient based on the error histogram, and use it on top of a Hamming distance to form our hierarchical loss with
\begin{equation}\label{eq_w}
    w_{j}(t) =\exp(\sigma \cdot  min\{H_j(t),0.5 - H_j(t)\}),
\end{equation}
where the function $w$ uses an exponential term to softly define a weight for $n$-th bit at training step $t$, $\sigma$ is a constant. All object pixels in the mini-batch share the same weighting coefficients. We normalize the weights across all bits. We then define our hierarchical loss based on the weighting function of active bits and Hamming distance as below:
\begin{equation}
    L_{hier} = \sum_{j=1}^{d}{w_j \cdot \text{$Hamm$} (b_j,\hat{p_j})}.
\end{equation}
With this loss we focus mainly on active bits which automatically change from coarse to fine during training. \\

\textbf{Total loss to train the CNN.} We weight the $L_{mask}$ and $L_{hier}$ with a hyper-parameter $\alpha$ ($\alpha$ set as $0$ for pixels predicted as background), the total per pixel loss can be mathematically expressed as
\begin{equation}
    L_{total} = L_{mask} + \alpha \cdot L_{hier}.
\end{equation}

\subsection{Pose estimation}\label{pose}
In previous sections, we discussed how to generate our descriptor and learn to predict them using a fully convolutional neural network. Now we incorporate the predicted code and visible mask and the reference 3D model encoding to match correspondences.
Different from common dense correspondences such as \cite{park2019pix2pose,wang2019normalized,wang2021gdr}, this compact representation also enables a bijective correspondence between the surface vertices and the descriptor space. That means, unlike the regressed 3D point which can be off the object surface, our estimated 3D correspondences always refers to a vertex on the object model, which eases the matching stage for the pose solver. For the matching, we use a look-up table that extracts the corresponding 2D and 3D points. Following that, we use Progressive-X \cite{barath2019progressive} solver to calculate the rotation $R$ and translation $t$.
\section{Experiments}
In this section, we firstly introduce the implementation details, the datasets and metrics used for the evaluation. Subsequently, we present ablation study experiments on the LM-O~\cite{brachmann2016uncertainty} dataset. Finally, we compare our experimental results with state of the art methods on the LM-O~\cite{brachmann2016uncertainty} and YCB-V~\cite{xiang2017posecnn} datasets. Please refer to supplementary materials for more qualitative results.

\subsection{Experiments Setup}\label{sec:Experiments_Setup}
\textbf{Implementation Details.} In order to have the same number of classes as DPOD~\cite{Zakharov2019DPOD6P} ($K=256^2$),  we firstly upsample the mesh by subdivision of each face using the edge's midpoint~\cite{chen2012general} until the mesh has more than $256^2$ vertices. Subsequently, we group the 3D vertices of the object model as we described in Sec.~\ref{sec:SurfaceEncoding} with $r=2$ and $d=16$. After several iterations of the grouping operation, a group could contain fewer points than 2 and cannot be grouped further. To avoid this, we modified the k-means++ clustering algorithm~\cite{arthur2006k}, to force both output groups of points to have equal sizes. 
  
We modified Deeplabv3~\cite{chen2017rethinking} by adding skip connections and used Resnet34~\cite{he2016deep} as the backbone. The input ROI is resized to the shape of 256$\times$256$\times$3, and the CNN output has a height and width of 128. We applied the same dynamic zoom-in strategy as CDPN~\cite{li2019cdpn} to generate the noisy ROI for the training. The parameter $\lambda$ used in the histogram is 0.05. The parameter $\sigma$ used in the hierarchical loss is 0.5, and $\alpha$ has been set as 3 to balance the training for mask and vertex code prediction. The CNN has been trained $380k$ steps using the Adam optimizer~\cite{kingma2014adam} with a batch size of 32 and a fixed learning rate of 2e-4. During the inference stage, we utilize the detected bounding box with Faster R-CNN~\cite{ren2015faster} and FCOS~\cite{tian2019fcos} provided by CDPNv2~\cite{li2019cdpn}. If not specified, we used detected bounding box from Faster R-CNN in the ablation study.

Additionally, by changing any bit in the vertex code, the code refers to another 3D point, possibly even to a vertex on the other side of the object. To maintain the topology presented with the ground truth correspondence map, we disabled the interpolation during the rendering when we generated the ground truth. The resizing of ground truth is also done with nearest neighbourhood interpolation in the training stage.

\textbf{Datasets.}
The reported recall rate in LM~\cite{hinterstoisser2012} dataset has lately been higher than $95\%$ and quite saturated, therefore we focus on the more challenging LM-O~\cite{brachmann2016uncertainty} and YCB-V~\cite{xiang2017posecnn} dataset in this paper. LM-O consist of 1214 images and is only used as test images. LM-O annotated 8 objects poses in the images under partial occlusion, making pose estimation more challenging. About $1.2k$ images per object in LM are used as the real training images for LM-O. Compared to LM-O, YCB-V is a large dataset containing 21 objects.
Although YCB-V provides more real training images, the objects are strongly occluded in the scene, and many of the objects are geometrically symmetric. 

Since the LM-O dataset includes only a limited number of training images, \cite{kehl2017ssd,peng2019pvnet} additionally render a large number of synthetic images for training. However, due to the domain gap between the synthetic and real images, the performance of the methods also heavily depends on the domain randomization and domain adaptation technique~\cite{sundermeyer2018implicit,zakharov2019deceptionnet}. As the physically-based rendering (pbr) training images~\cite{denninger2019blenderproc} for both datasets are publicly accessible now, using pbr images to support the training can help us focus on the pose estimation CNN itself.  We use the pbr images together with the real images for the training in the same manner as \cite{hodan2020epos,wang2021gdr,di2021so}.

\textbf{Error Metrics.}
We selected the ADD(-S) error metric as the most commonly used metric for the 6DoF pose estimation task. This metric calculates the average distance of model points projected to the camera domain using the predicted pose to the same model points projected using the ground truth pose.  For symmetric objects, the metric matches the closest model points projected with the ground truth pose instead of the same model point. In all the experiments in this paper, if ADD(-S) error is smaller than 10\% (most commonly used threshold) of the object diameter, the predicted pose is considered to be correct 
For YCB-V, we also reported the AUC
(area under curve) of ADD(-S)  with a maximum threshold of 10 cm \cite{xiang2017posecnn}.

\subsection{Ablation Study on LM-O}\label{sec:ablation_study}

\begin{figure*}
    \centering
    \includegraphics[width=\linewidth]{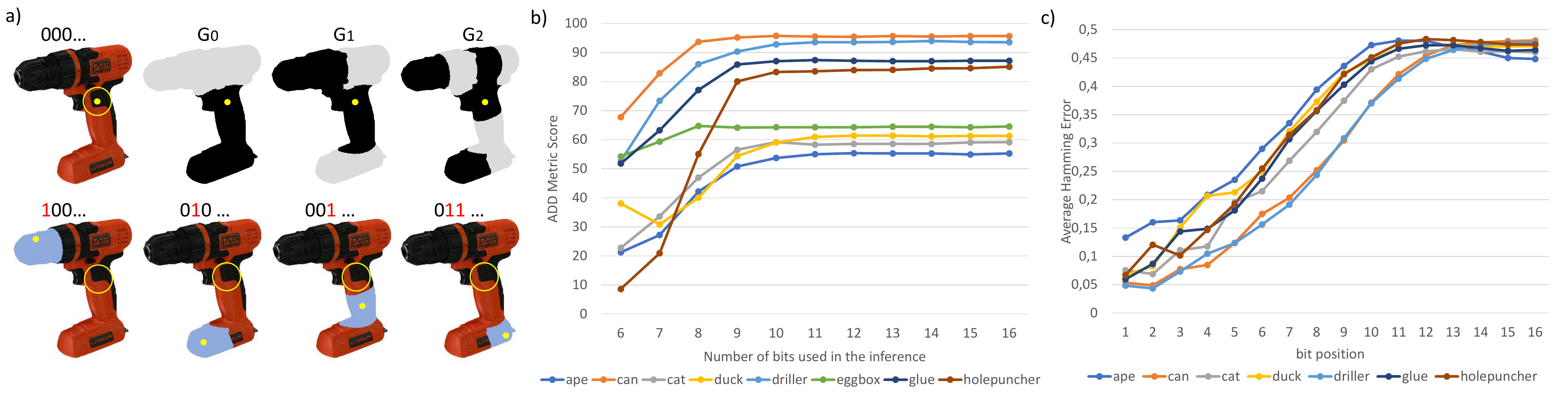}
    \caption{a) In the first row, a yellow dot has the ground truth binary vertex code beginning with {000}, and the yellow circle refers to the neighborhood vertex of this yellow dot. If the vertex code has been predicted as {100} (the first bit is wrong, marked as red in the figure) during the inference stage, the estimated yellow dot lies somewhere on the head of the drill  (marked in blue). The estimated 3D vertex is far away from its original neighborhood and can be easily found by checking the spatial coherency. We show four similar cases in this figure. b) We calculated ADD pose metrics only on the first $j$ bits of the predicted code to build the 2D-3D correspondence. Here we observe from which bit the predictions are stable. c) We present the average error rate at different bit positions on the LM-O dataset\cite{brachmann2016uncertainty}.}
    \label{fig:ablation_study}
\end{figure*}

In this section, we present the results of several ablation studies as follows:

\textbf{Length of Binary Vertex Code.}
The object 3D surface is encoded through iterative k-means++ clustering until the size of the segmented cluster is small enough so that we can map the vertex code to the centroid of each cluster. We used the same total number of classes as DPOD~\cite{Zakharov2019DPOD6P}, which means each binary vertex code has 16 bits. However, if the objects are small or the distance of the object to the camera is too large, different clusters in the fine level could be rendered into the same pixel when we generate the ground truth data. This makes the binary code in the fine levels (the last few bits) redundant. Due to the distance variation of the object to the camera, we can not determine which bits are redundant. 

In this ablation study, we research which bits are the redundant bits. The models are trained without the hierarchical training strategy, and we use Progressive-X~\cite{barath2019progressive} to solve the pose. We ignore the last few bits of the predicted binary code in the inference stage. The new binary vertex code with fewer bits refers to a larger point cloud group (see Fig.~\ref{fig:overview} left). We calculated the new centroid of the group and reassigned the centroid as the corresponding 3D points for the binary vertex code with fewer bits. From Fig.~\ref{fig:ablation_study} b) we can see that using 10-bit code is already sufficient to yield an accurate prediction for the objects in LM-O, indicating the last 6 bits are redundant for those objects.
The results fluctuated a bit when we applied the redundant bits, which indicates that for some objects the best results is achieved by not using the full 16-bit code. However, in the following experiment, we always report the results with the full predicted vertex code.

\textbf{Radix used in Vertex Code.}  The number of the clusters in each iteration decides the radix of the generated vertex code that describes the 3D vertex. Since our CNN predicts the vertex code, it is meaningful to compare which radix in vertex code suits the representation better. We do not need to generate all the vertex codes used in this ablation study from scratch. More specifically, by merging every $log_2 r$ bit of a vertex code, we get a code with a radix $r$. For instance, a vertex with a binary code \{11111110 11111111\} can be transformed to \{254 255\} using 256 as the radix. We will get exactly the same code for this vertex if we split the object into 256 groups and split each group again into 256 groups. We use a fix $w_j=1$ (see eq. \ref{eq_w}) for all positions so the loss is essentially a binary cross entropy when $r=2$, and cross-entropy loss for other radixes. 

We present the comparison results in Tab.~\ref{tab:ablation_study_radix}. If RANSAC/PnP is used to solve the pose, the results with different radices are quite similar. There is no clear indication whether using the small or large radix is better. If we switch the pose solver to Progressive-X~\cite{barath2019progressive}, the code with small radix improves the most and yields the best accuracy. 
Progressive-X solver includes a spatial coherence filter that checks neighboring 3D points with respect to its assigned 2D correspondences based on label cost energy minimization as introduced in \cite{delong2012fast}. 
It can therefore deal particularly well with the type of outliers, from our method as we mentioned in Fig.~\ref{fig:ablation_study}. We show in the Fig.~\ref{fig:ablation_study} a), if the CNN predicts the first few bits wrong for the yellow dot, the estimated corresponding 3D points is far away from the ground truth position and totally incoherent with its original neighbourhood. Assuming that most predictions are correct, most neighbourhood vertices are posed within the yellow circle in the figure. In this case, this false estimated 3D corresponding can be easily filtered by calculating the coherency with its neighbourhood. This spatial coherency filter can not detect outliers well if the wrong prediction is in the last few bits, and the same for 256 as radix, as divide the vertices into 256 groups is already a fine grouping. Nevertheless, the Fig.~\ref{fig:ablation_study} b) already shows that the last few bits do not affect the solved pose. So we argue that the binary vertex code suits this task the best. Moreover, the prediction of binary codes requires the least RAM in GPU, as we discussed in Sec.~\ref{sec:SurfaceEncoding}.

\textbf{Effectiveness of Hierarchical Training.}
According to the first ablation study, the last few bits are redundant and may not be trainable (see Fig.~\ref{fig:ablation_study}c)). During the training, we can recognize redundant bits based on the error histogram and focus on the decisive bits as described in Sec.~\ref{Learning}. Tab.~\ref{tab:ablation_study_network} shows that the results are further improved by our proposed hierarchical training. 

\textbf{Influence of 2D detection.}
The CNN estimates the pose with the cropped ROI from the detected bounding box. The object pose estimation is meaningless with a false-positive detection, also the pose is not even estimated in the case of false-negative detection. By leveraging the detected bounding box with FCOS~\cite{tian2019fcos} instead of the one from Faster R-CNN~\cite{ren2015faster}, the recall rate improved 1.05\%.

\begin{table}
  \centering
  \begin{tabular}{@{}c|c|c@{}}
    \toprule
     Method & RANSAC/ Pnp~\cite{epnp_2009} &  Progressive-X  \cite{barath2019progressive}\\
    \midrule
     2 as radix &73.06 &\textbf{75.23 (+2.17)}  \\  
     4 as radix &72.94 &74.59 (+1.65) \\
     16 as radix &73.04 &74.98 (+1.94)\\
     256 as radix &73.25 &74.52 (+1.27)\\
    \bottomrule
  \end{tabular}
  \caption{\textbf{Ablation study on LM-O~\cite{brachmann2016uncertainty}.} We tested the use of different radices to encode the vertices, and using different solvers to calculate the pose. The results are presented in terms of average recall of ADD(-S) in \%.}
  \label{tab:ablation_study_radix}
\end{table}

\begin{table}
  \centering
  \begin{tabular}{@{}c|c@{}}
    \toprule
     Method  & ADD\\
    \midrule
    \multicolumn{1}{l|}{2 as radix}  &75.23  \\  

    \hline
     \multicolumn{1}{l|}{2 as radix + Hierarchical Learning}  &75.86\\
    \hline
     \multicolumn{1}{l|}{\makecell[l]{2 as radix + Hierarchical Learning  \\
     + Faster R-CNN~\cite{ren2015faster} $\rightarrow$ FCOS~\cite{tian2019fcos}} }  &\textbf{76.91}\\
    \bottomrule
  \end{tabular}
  \caption{\textbf{Ablation study on LM-O\cite{brachmann2016uncertainty}.} We compare the result with and w/o applying our hierarchical loss, as well as the impact of the prior object detector. The results are presented with average recall of ADD(-S) in \%.}
  \label{tab:ablation_study_network}
\end{table}

\begin{table*}
  \centering
  \begin{tabular}{@{}c|c|c|c|c|c|c|c@{}}
    \toprule 
     \multirow{2}{*}{Method} & \multicolumn{5}{c|}{RGB Input} & \multicolumn{2}{c}{RGB-D Input} \\ 
     \cline{2-8}
     & HybridPose~\cite{song2020hybridpose} & RePose~\cite{iwase2021repose} &  GDR-Net~\cite{wang2021gdr} &  SO-Pose~\cite{di2021so} & \textbf{Ours} &PR-GCN~\cite{zhou2021pr} &FFB6D~\cite{he2021ffb6d}\\
    \midrule
     ape & 20.9 &  31.1 &  46.8 &  48.4 & \textbf{57.9} &  40.2 &  47.2  \\
     can & 75.3 &  80.0 &  90.8 &  85.8 & \textbf{95.0} &  76.2 &  85.2  \\
     cat & 24.9 &  25.6 &  40.5 &  32.7 & \textbf{60.6} &  57.0 &  45.7  \\
     driller & 70.2 &  73.1 &  82.6 &  77.4 & \textbf{94.8} &  82.3 &  81.4  \\
     duck & 27.9 &  43.0 &  46.9 &  48.9 & \textbf{64.5} &  30.0 &  53.9  \\
     eggbox* & 52.4 &  51.7 &  54.2 &  52.4 & \textbf{70.9} &  68.2 &  70.2  \\
     glue* & 53.8 &  54.3 &  75.8 &  78.3 & \textbf{88.7} &  67.0 &  60.1  \\
     holepuncher & 54.2 &  53.6 &  60.1 &  75.3 & 83.0 &  \textbf{97.2} &  85.9  \\
     \hline
     mean & 47.5 &  51.6 &  62.2 &  62.3 & \textbf{76.9} &  65 & 66.2   \\
    \bottomrule
  \end{tabular}
  \caption{\textbf{Comparison with State of the Art on LM-O\cite{brachmann2016uncertainty}}. We report the Recall of ADD(-S) in \% and compare with state of the art. (*) denotes symmetric objects.}
  \label{tab:lmo_results_table_}
\end{table*}
\subsection{Comparison to State of the Art}\label{sec:Comparison_with_State_of_the_Art}
We use 2 as radix, i.e. binary vertex code and apply the hierarchical training strategy and Progressive-X pose solver~\cite{barath2019progressive} in our proposed ZebraPose to compare to state of the art on LM-O~\cite{brachmann2016uncertainty} and YCB-V~\cite{xiang2017posecnn} datasets. The detected bounding box of FCOS~\cite{tian2019fcos} detector are provided by CDPNv2~\cite{li2019cdpn}.

\textbf{Results on LM-O.} We report the recall of ADD(-S) metric in Tab.~\ref{tab:lmo_results_table_}. We ordered the methods according to the input modality. HybridPose~\cite{song2020hybridpose} and RePose~\cite{iwase2021repose} have been trained with synthetic and real images. GDR-Net~\cite{wang2021gdr} also reported their recall of 53\% when trained with synthetic and real images. Therefore, GDR-Net outperforms HybridPose and RePose. In our Tab.\ref{tab:lmo_results_table_}, we report the best results that GDR-Net and SO-Pose~\cite{di2021so} presented, which are also trained with pbr and real images. GDR-Net used faster R-CNN~\cite{ren2015faster} as the detector, ZebraPose yields a recall of 75.86\% with faster R-CNN (see Tab.~\ref{tab:ablation_study_network}), which can be seen as a  more fair comparison with GDR-Net.

To summarize, our ZebraPose outperforms state of the art RGB based methods with a large margin on this dataset. Additionally, we found that our ZebraPose also outperforms state of the art RGB-D based methods~\cite{zhou2021pr,he2021ffb6d}. Most objects in the LM-O dataset are texture-less, meaning that RGB-D based methods should have more advantage in feature extraction on the objects with the help of depth image. Even in this case, our results still exceed theirs.

\textbf{Results on YCB-V.}
We compare ZebraPose with other approaches in the YCB-V dataset in Tab.~\ref{tab:ycbv_results_table}. The AUC reported in Tab.~\ref{tab:ycbv_results_table} has been calculated using all-points interpolation. Tab.~\ref{tab:ycbv_results_table} shows that ZebraPose is still better than state of the art w.r.t. ADD(-S) and AUC of ADD(-S) metrics and comparable to them w.r.t. the AUC of ADD-S metric. 

\begin{table}
  \centering
  \begin{tabular}{@{}l|c|c|c@{}}
    \toprule
     Method & ADD(-S) & \makecell{AUC of\\ADD-S} & \makecell{AUC of\\ADD(-S)} \\
    \midrule
    SegDriven\cite{hu2019segmentation} & 39.0 &  - &  -  \\
    SingleStage\cite{hu2020single} & 53.9 &  - &  - \\
    CosyPose~\cite{labbe2020cosypose} & - &  89.8 &  84.5 \\
    RePose~\cite{iwase2021repose} & 62.1 &  88.5 &  82.0 \\
    GDR-Net~\cite{wang2021gdr} & 60.1 &  \textbf{91.6} &  84.4 \\
    SO-Pose~\cite{di2021so} & 56.8 &  90.9 &  83.9 \\
    Ours & \textbf{80.5} &  90.1 &  \textbf{85.3}  \\
    \bottomrule
  \end{tabular}
  \caption{\textbf{Comparison with State of the Art on YCB-V\cite{xiang2017posecnn}}. We compare our ZebraPose with state of the art w.r.t ADD(-S), AUC of ADD(-S) and AUC of ADD-S in \%. (-) denotes results missing from the original paper.}
  \label{tab:ycbv_results_table}
\end{table}

\subsection{Runtime Analysis}
We tested the runtime on a desktop with an Intel 3.50GHz CPU and an Nvidia 2080Ti GPU. The CNN runtime plus the time to build the 2D-3D correspondence is about 52 ms. The FCOS detector\cite{tian2019fcos} takes 55 ms. RANSAC/PnP~\cite{epnp_2009} needs only 4 ms to solve the pose, while Progressive-X~\cite{barath2019progressive} requires 150 ms to obtain the pose. So for ZebraPose used in Sec.\ref{sec:Comparison_with_State_of_the_Art}, it totally needs about 250 ms to estimate the object pose. If we use RANSAC/PnP to solve the pose, the runtime reduces to 110 ms, while with about 2.6\% recall drop on LM-O dataset.
\section{Conclusion}
In this work, we proposed a novel coarse to fine surface encoding technique to provide 2D-3D correspondences for 6DoF object pose estimation. We also designed a specific hierarchical training strategy that maximizes the prediction accuracy for our proposed binary vertex code. Solving the object pose using a PnP solver based on our vertex code surpasses the state of the art on different benchmarks, proving our approach's effectiveness. In the future, we would like to extend our vertex code solution to the problem of category-level object pose~\cite{li2020category}. 

\section*{Acknowledgements} 
This work was partially funded by the \emph{Federal Ministry of Education and Research} of the Federal Republic of Germany (BMBF), under grant agreements 16SV8732 (GreifbAR) and 01IW21001 (DECODE). We are thankful to Rene Schuster, Fangwen Shu, Yaxu Xie and Ghazal Ghazaei for proofreading the paper.
{\small
\bibliographystyle{ieee_fullname}
\bibliography{egbib}
}

\newpage
\section{Supplementary Material }


\subsection{Hyper-parameters in the Pose Solver}
For RANSAC/PnP~\cite{lepetit2009epnp}, we set the threshold value for reprojection error as 2 pixels, and execute 150 iterations. For Progressive-X~\cite{barath2019progressive}, we 
also set the threshold value for the reprojection error as 2 pixels, and execute 400 iterations. The additional parameters for Progressive-X are "neighborhood\_ball\_radius=20", "spatial\_coherence\_weight=0.1", "maximum\_tanimoto\_similarity=0.9".

\subsection{BOP Challenge}
We submitted the results on 4 datasets of the BOP challenge and will test our method on the rest 3 datasets. The results are online in \href{/https://bop.felk.cvut.cz/leaderboards/}{BOP Leaderboards} with the submission name "zebrapose".

\subsection{YCB-V Evaluation per Object}
We present a more detailed result on the YCB-V dataset~\cite{xiang2017posecnn} in Tab.~\ref{tab:ycbv_full_results_ADD} and Tab.~\ref{tab:ycbv_full_results_AUC}. As the Tab.~\ref{tab:ycbv_full_results_ADD} shows, in the evaluation of the estimate pose w.r.t ADD(-S) metric, we show major improvement over the state of the art.

In Tab.~\ref{tab:ycbv_full_results_AUC}, we carefully calculated the AUC with all-points interpolation algorithm with the maximum threshold of 10 cm. If we calculate the AUC with 11-points interpolation, we will reach AUC of ADD-S of 94\%, and AUC of ADD(-S) of 89.8\%. 

\begin{table*}
  \centering
  \begin{tabular}{@{}l|c|c|c|c|c@{}}
    \toprule 
     Method & SegDriven~\cite{hu2019segmentation} & Single-Stage~\cite{hu2020single} & RePose~\cite{iwase2021repose} &  GDR-Net~\cite{wang2021gdr}  & \textbf{Ours}\\
    \midrule
     002\_master\_chef\_can & 33.0 &  - &  - &  41.5 &   \textbf{62.6}  \\
     003\_cracker\_box      & 44.6 &  - &  - &  83.2  &  \textbf{98.5}  \\
     004\_sugar\_box        & 75.6 &  - &  - &  91.5  &  \textbf{96.3}  \\
     005\_tomato\_soup\_can & 40.8 &  - &  - &  65.9  &  \textbf{80.5}   \\
     006\_mustard\_bottle   & 70.6 &  - &  - &  90.2  &  \textbf{100.0}   \\
     007\_tuna\_fish\_can   & 18.1 &  - &  - &  44.2  &  \textbf{70.5}  \\
     008\_pudding\_box      & 12.2 &  - &  - &  2.8  &  \textbf{99.5} \\
     009\_gelatin\_box      & 59.4 &  - &  - &  61.7  &  \textbf{97.2}  \\
     010\_potted\_meat\_can & 33.3 &  - &  - &  64.9  &  \textbf{76.9}  \\
     011\_banana            & 16.6 &  - &  - &  64.1  & \textbf{71.2} \\
     019\_pitcher\_base     & 90.0 &  - &  - &  99.0  & \textbf{100.0}  \\
     021\_bleach\_cleanser  & 70.9 &  - &  - &  73.8  & \textbf{75.9}  \\
     024\_bowl*             & 30.5 &  - &  - &  \textbf{37.7}  & 18.5  \\
     025\_mug               & 40.7 &  - &  - &  61.5  &  \textbf{77.5}  \\
     035\_power\_drill      & 63.5 &  - &  - &  78.5  &  \textbf{97.4}  \\
     036\_wood\_block*      & 27.7 &  - &  - &  59.5  &  \textbf{87.6}  \\
     037\_scissors          & 17.1 &  - &  - &  3.9  &  \textbf{71.8}  \\
     040\_large\_marker     & 4.8 &  - &  - &  7.4  &  \textbf{23.3}  \\
     051\_large\_clamp*     & 25.6 &  - &  - &  69.8  &  \textbf{87.6}  \\
     052\_extra\_large\_clamp* & 8.8 &  - &  - &  90.0  &  \textbf{98.0}  \\
     061\_foam\_brick*          & 34.7 &  - &  - &  71.9  &  \textbf{99.3} \\
     \hline
     mean & 39.0 &  53.9 &  62.1 &  60.1 &  \textbf{80.5}   \\
    \bottomrule
  \end{tabular}
  \caption{\textbf{Comparison with State of the Art on YCB-V}. We report the Average Recall of ADD(-S) in \% and compare with state of the art. (*) denotes symmetric objects, (-) denotes the results missing from the original paper.}
  \label{tab:ycbv_full_results_ADD}
\end{table*}

\begin{table*}
  \centering
  \begin{tabular}{@{}l|c|c|c|c|c|c|c|c@{}}
    \toprule 
     Method & \multicolumn{2}{c|}{PoseCNN~\cite{xiang2017posecnn}} & \multicolumn{2}{c|}{CosyPose\cite{labbe2020cosypose}} &  \multicolumn{2}{c|}{GDR-Net~\cite{wang2021gdr}} &  \multicolumn{2}{c}{\textbf{Ours}}\\
    \midrule
    Metric & \makecell{AUC of \\ADD-S} & \makecell{AUC of \\ADD(-S)} &\makecell{AUC of \\ADD-S} & \makecell{AUC of \\ADD(-S)} & \makecell{AUC of \\ADD-S} & \makecell{AUC of \\ADD(-S)} & \makecell{AUC of \\ADD-S} & \makecell{AUC of \\ADD(-S)} \\ 
    \midrule
     002\_master\_chef\_can & 84.0 &  50.9 &  - &  - & 96.3 &  65.2 & 93.7  & 75.4\\
     003\_cracker\_box      & 76.9 &  51.7 &  - &  - & 97.0 &  88.8 & 93.0  & 87.8\\
     004\_sugar\_box        & 84.3 &  68.6 &  - &  - & 98.9 &  95.0 & 95.1  & 90.9\\
     005\_tomato\_soup\_can & 80.9 &  66.0 &  - &  - & 96.5 &  91.9 & 94.4  &  90.1\\
     006\_mustard\_bottle   & 90.2 &  79.9 &  - &  - & 100 &  92.8  & 96.0  & 92.6\\
     007\_tuna\_fish\_can   & 87.9 &  70.4 &  - &  - & 99.4 &  94.2 & 96.9  & 92.6\\
     008\_pudding\_box      & 79.0 &  62.9 &  - &  - & 64.6 &  44.7 & 97.2  & 95.3\\
     009\_gelatin\_box      & 87.1 &  75.2 &  - &  - & 97.1 &  92.5  & 96.8  & 94.8\\
     010\_potted\_meat\_can & 78.5 &  59.6 &  - &  - & 86.0 &  80.2  & 91.7  & 83.6\\
     011\_banana            & 85.9 &  72.3 &  - &  - & 96.3 & 85.8  & 92.6  & 84.6\\
     019\_pitcher\_base     & 76.8 &  52.5 &  - &  - & 99.9 & 98.5  & 96.4  & 93.4\\
     021\_bleach\_cleanser  & 71.9 &  50.5 &  - &  - & 94.2 & 84.3  & 89.5  & 80.0\\
     024\_bowl*             & 69.7 &  69.7 &  - &  - & 85.7 & 85.7   & 37.1  & 37.1\\
     025\_mug               & 78.0 &  57.7 &  - &  - & 99.6 &  94.0  & 96.1  & 90.8\\
     035\_power\_drill      & 72.8 &  55.1 &  - &  - & 97.5 & 90.1  & 95.0  & 89.7\\
     036\_wood\_block*      & 65.8 &  65.8 &  - &  - & 82.5 &  82.5  & 84.5  & 84.5\\
     037\_scissors          & 56.2 &  35.8 &  - &  - & 63.8 & 49.5  & 92.5  & 84.5\\
     040\_large\_marker     & 71.4 &  58.0 &  - &  - & 88.0 & 76.1  & 80.4  & 69.5\\
     051\_large\_clamp*     & 49.9 &  49.9 &  - &  - & 89.3 & 89.3  & 85.6  & 85.6\\
     052\_extra\_large\_clamp* & 47.0 &  47.0 &  - &  - & 93.5 & 93.5 & 92.5  & 92.5\\
     061\_foam\_brick*          & 87.8 &  87.8 &  - &  - & 96.9 & 96.9 & 95.3  & 95.3\\
     \hline
     mean & 75.9 &  61.3 &  89.8 &  84.5 & 91.6 &  84.3  &90.1 &85.3 \\
    \bottomrule
  \end{tabular}
  \caption{\textbf{Comparison with State of the Art on YCB-V}. We report the Average Recall w.r.t AUC of ADD(-S) and AUC of ADD-S in \% and compare with state of the art. (*) denotes symmetric objects, (-) denotes the results missing from the original paper.}
  \label{tab:ycbv_full_results_AUC}
\end{table*}

\subsection{Qualitative Results}
\subsubsection{Vertex Code Prediction LM-O}
\begin{figure*}
    \centering
    \includegraphics[width=0.95\linewidth]{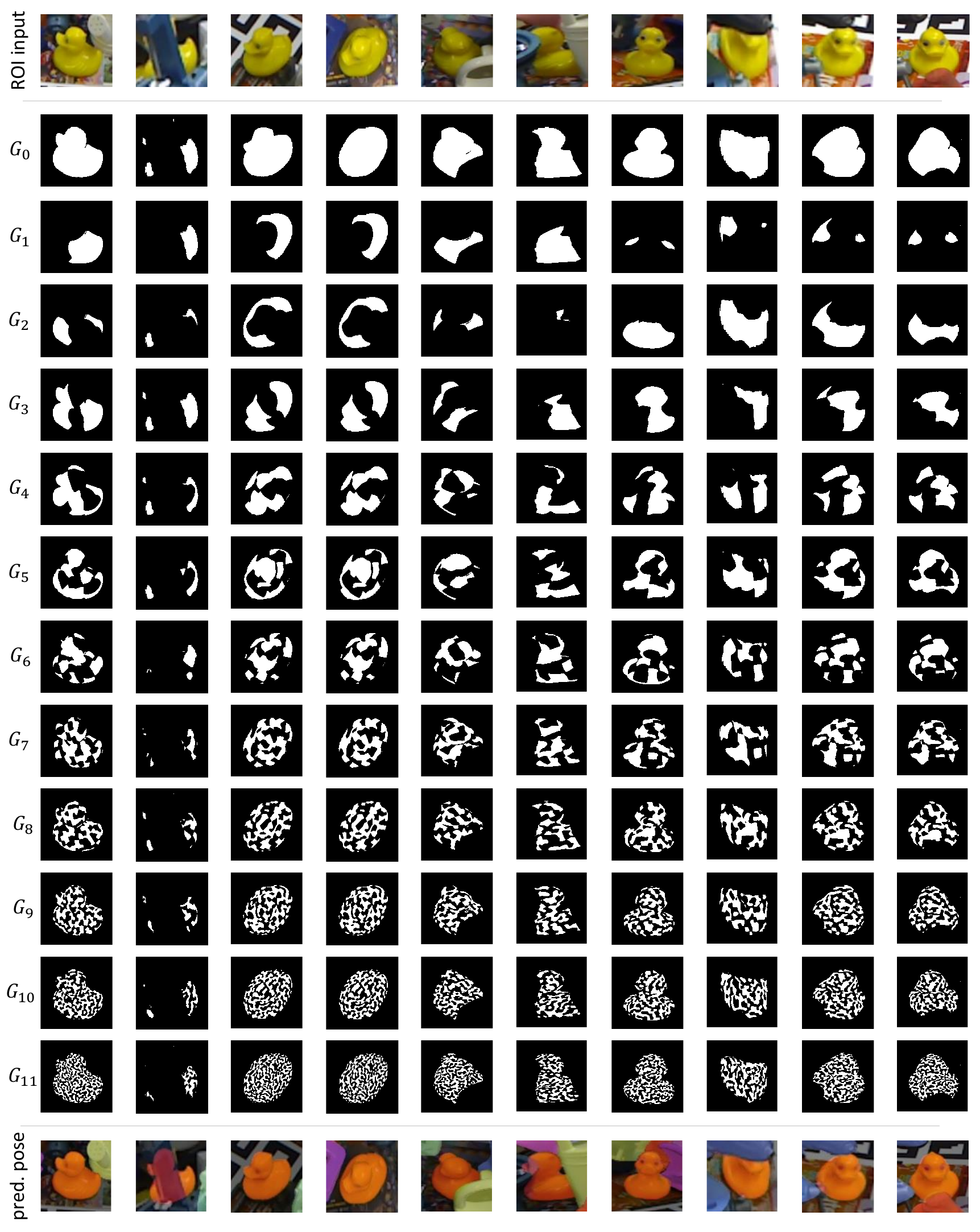}
    \caption{We visualized the predicted binary code of the "duck" in LM-O dataset\cite{brachmann2016uncertainty} with a few examples. Due to the size limits, we only show the predicted binary code till the 11-th bit. We set the colour of the object model as red and render the object with the predicted pose on the top of the input ROI. We can see that the rendered object overlaps the object in the image quite well.}
    \label{fig:supp_lmo_binary_code}
\end{figure*}

We visualized the predicted binary code of the "duck" object in LM-O dataset\cite{brachmann2016uncertainty} with a few examples in Fig.~\ref{fig:supp_lmo_binary_code}. Due to the size limits, we only show the predicted binary code till the 11-th bits. We render the object with the predicted pose on top of the original input ROI. To make the predicted pose more visible in the figure, we set the colour of the object model as red just for this figure. So the duck appears with the orange colour (red + yellow) in the last row. We can see that the rendered object overlapped the object in the original image quite well, indicating that our predicted pose is very accurate.

\subsubsection{Pose Prediction LM-O}
\begin{figure*}
    \centering
    \includegraphics[width=0.95\linewidth]{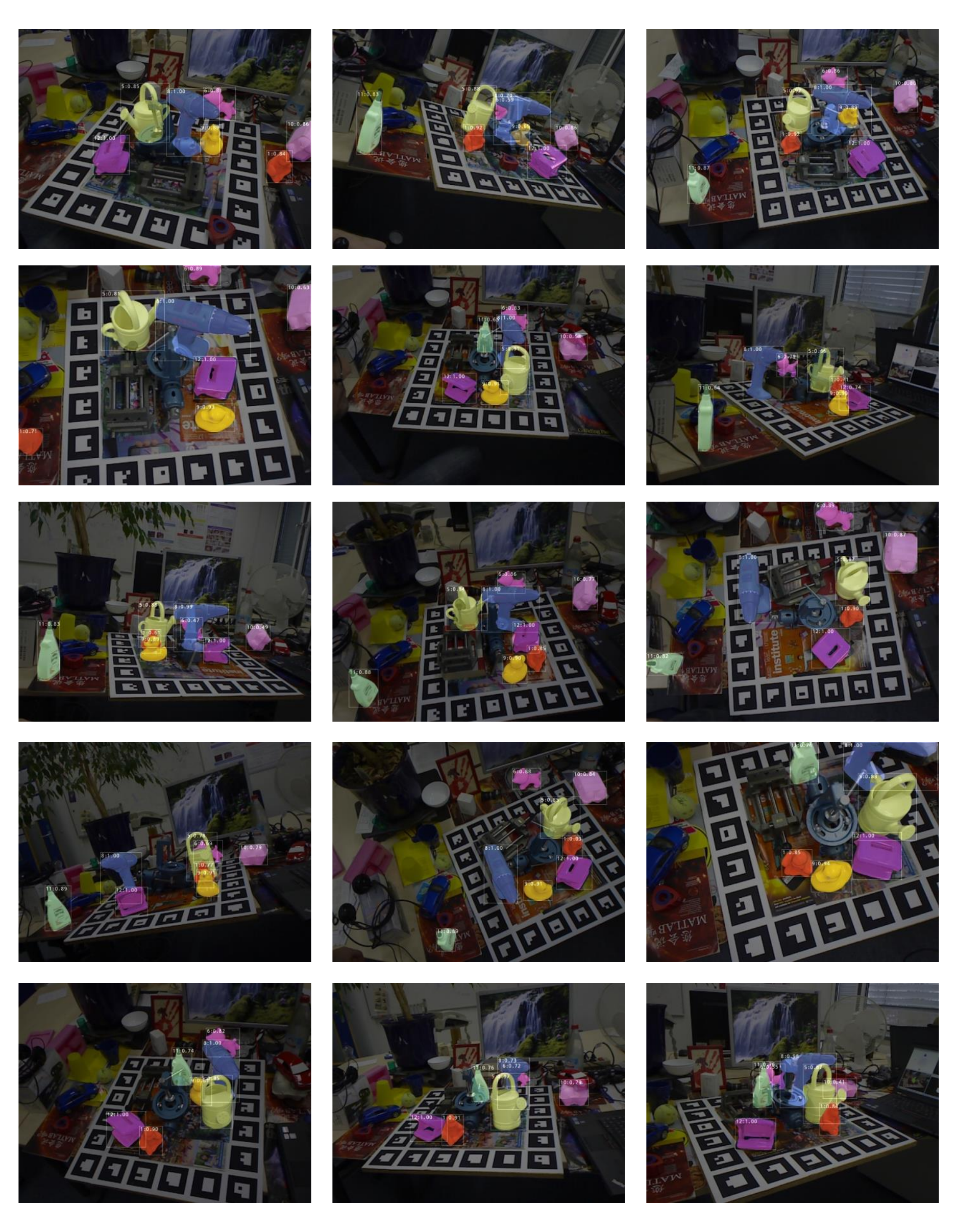}
    \caption{\textbf{Qualitative Results on LM-O\cite{brachmann2016uncertainty}}: We render the objects with estimated pose on top of the original images. The presented confidence score are from the 2D object detection with FCOS detector~\cite{tian2019fcos}.}
    \label{fig:supp_lmo_scenes}
\end{figure*}

Qualitative Results on LM-O\cite{brachmann2016uncertainty} can be found in Fig.~\ref{fig:supp_lmo_scenes}. We render the objects with estimated pose on top of the original images. The presented confidence scores are from the 2D object detection with FCOS detector~\cite{tian2019fcos}.
\subsubsection{Pose Prediction YCB-V}
\begin{figure*}
    \centering
    \includegraphics[width=0.95\linewidth]{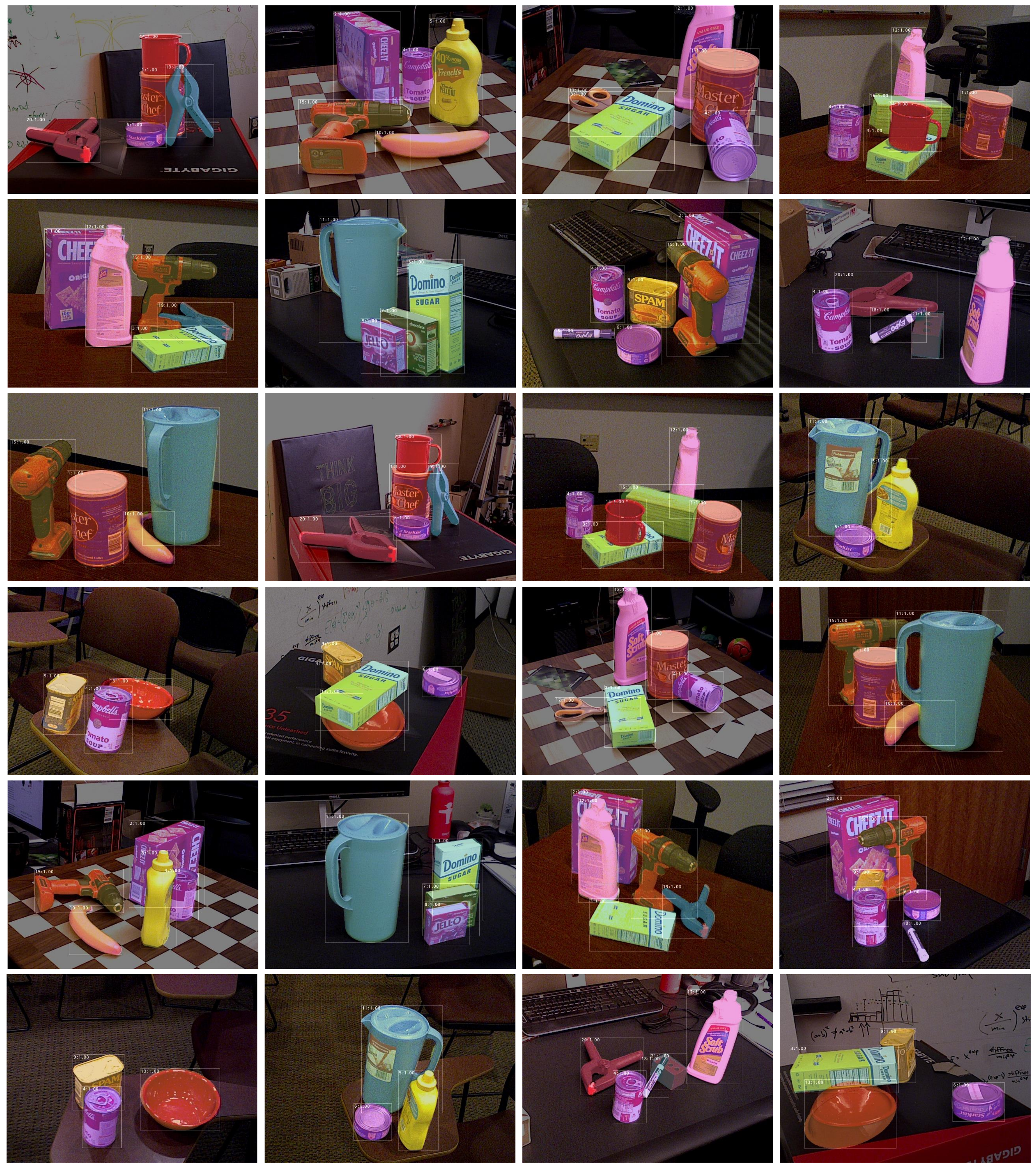}
    \caption{\textbf{Qualitative Results on YCB-V\cite{xiang2017posecnn}}: We render the objects with estimated pose on top of the original images. The presented confidence score are from the 2D object detection with FCOS detector~\cite{tian2019fcos}.}
    \label{fig:supp_ycbv_scenes}
\end{figure*}

Qualitative Results on YCB-V\cite{xiang2017posecnn} are available in Fig.~\ref{fig:supp_ycbv_scenes}. We render the objects with estimated pose on top of the original images. The presented confidence scores are from the 2D object detection with FCOS detector~\cite{tian2019fcos}.

\clearpage

\end{document}